\pdfoutput=1
\documentclass[11pt]{article}

\usepackage[]{ACL2023}

\usepackage{times}
\usepackage{fancyvrb}
\usepackage{varwidth}
\usepackage{latexsym}
\usepackage{listings}
\usepackage{xcolor}

\usepackage[most]{tcolorbox}

\definecolor{light-gray}{gray}{0.95}

\tcbset{on line, 
        boxsep=3pt,left=0pt,right=0pt,top=0pt,bottom=0pt,
        colframe=white,colback=light-gray,  
        highlight math style={enhanced}
        }

\definecolor{codegreen}{rgb}{0,0.6,0}
\definecolor{codegray}{rgb}{0.5,0.5,0.5}
\definecolor{codepurple}{rgb}{0.58,0,0.82}
\definecolor{backcolour}{rgb}{0.95,0.95,0.92}

\lstdefinestyle{mystyle}{
    backgroundcolor=\color{backcolour},   
    commentstyle=\color{codegreen},
    keywordstyle=\color{magenta},
    numberstyle=\tiny\color{codegray},
    stringstyle=\color{codepurple},
    basicstyle=\ttfamily\footnotesize,
    breakatwhitespace=false,         
    breaklines=true,                 
    captionpos=b,                    
    keepspaces=true,                 
    numbers=left,                    
    numbersep=5pt,                  
    showspaces=false,                
    showstringspaces=false,
    showtabs=false,                  
    tabsize=2
}

\usepackage[T1]{fontenc}
\usepackage[utf8]{inputenc}

\usepackage{microtype}

\usepackage{inconsolata}

\usepackage{adjustbox}
\usepackage{wrapfig}
\usepackage{graphicx}
\usepackage{amssymb,amsmath,array}
\usepackage{pifont}
\usepackage{multirow,multicol,makecell}
\usepackage{url}
\usepackage{latexsym}
\usepackage{array}
\usepackage{todonotes}
\usepackage{subfigure} 
\usepackage{algorithm}
\usepackage{algorithmic}
\usepackage{makecell}
\usepackage{booktabs}
\usepackage{listings}
\usepackage{csquotes}
\usepackage{amsthm}
\usepackage{bm}
\usepackage{xcolor,colortbl}
\usepackage{xspace}
\usepackage{cleveref}
\crefformat{section}{\S#2#1#3}
\crefformat{subsection}{\S#2#1#3}

\title{When and What to Ask Through World States and Text Instructions \\ \textit{\large IGLU NLP Challenge Solution}}

\author{Zhengxiang Shi\thanks{~~Equal Contribution} ~ Jerome Ramos$^*$ ~ To Eun Kim ~ Xi Wang \\ 
\textbf{Hossein A. Rahmani} ~ \textbf{Aldo Lipani}  \\
\texttt{\{zhengxiang.shi.19, jerome.ramos.20, to.kim.17, xi-wang} \\
\texttt{hossein.rahmani.22, aldo.lipani\}@ucl.ac.uk}\\
Web Intelligence Group \\
        University College London, London, United Kingdom}

\begin{document}
\maketitle

% \begin{abstract}
% This document is a supplement to the general instructions for *ACL authors. 
% \end{abstract}

\section{Introduction}
In collaborative tasks, effective communication is crucial for achieving joint goals \cite{10.1145/3539813.3545126}.
One such task is collaborative building \cite{narayan-chen-etal-2019-collaborative} where builders must communicate with each other to construct desired structures in a simulated environment such as \textsc{Minecraft}. We aim to develop an intelligent builder agent to build structures based on user input through dialogue. However, in collaborative building, builders may encounter situations that are difficult to interpret based on the available information and instructions, leading to ambiguity. In the NeurIPS 2022 Competition NLP Task \cite{kiseleva2022iglu}, we address two key research questions, with the goal of filling this gap: when should the agent ask for clarification, and what clarification questions should it ask? We move towards this target with two sub-tasks, a classification task and a ranking task. For the classification task, the goal is to determine whether the agent should ask for clarification based on the current world state and dialogue history. For the ranking task, the goal is to rank the relevant clarification questions from a pool of candidates. In this report, we briefly introduce our methods for the classification and ranking task. For the classification task, our model achieves an F1 score of 0.757, which placed the 3rd on the leaderboard. For the ranking task, our model achieves about 0.38 for Mean Reciprocal Rank by extending the traditional ranking model. Lastly, we discuss various neural approaches for the ranking task and future direction.

%%%%%%%%%%%%%%%%%%%%%%%%%%%
%     Classification      %
%%%%%%%%%%%%%%%%%%%%%%%%%%%
\section{Classification Task}
% \cite{shi-2022-learning}
In this section, we introduce the proposed builder model, built upon \citet{shi-2022-learning}, as shown in Figure~\ref{fig:model}, and then present the experimental results.

\subsection{Method}
The model comprises four major components: the \textit{utterance encoder}, the \textit{world state encoder}, the \textit{fusion module}, and the \textit{slot decoder}.
The utterance encoder (\cref{sec:text_encoder}) and world state encoder (\cref{sec:world_encoder}) learn to represent the dialogue context and the world state.  
These encoded representations are then fed into the fusion module (\cref{sec:fusin_module}) that learns contextualized embeddings for the grid world and textual tokens through the single and cross-modality modules. Finally, the learned world and text representations are mapped into scalar values for the binary classification (\cref{sec:decoder}).

\paragraph{Dialogue Context Encoder.}
\label{sec:text_encoder}
We add ``architect'' and ``builder'' annotations before each architect utterance $A_t$ and each builder utterance $B_t$ respectively. Then, the dialogue utterances are represented as 
$$D_t = \text{``architect''} A_t \oplus \text{``builder''} B_t$$ at the turn $t$, where $\oplus$ is the operation of sequence concatenation.% and $t=1,\dots,n$. 
The entire dialogue context is defined as:
\begin{align}
    \footnotesize
    H = D_1 \oplus D_2 \oplus \dots \oplus D_t
\end{align}
Given the dialogue context $H$, we truncate the tokens from the end of the dialogue context or pad them to a fixed length as inputs and then use the dialogue context encoder to encode utterance history into 
$U \in \mathbb{R}^{s \times d_w}$, where $d_w$ is the dimension of the word embedding and $s$ is the maximum number of tokens for a dialogue context. BERT~\cite{devlin2018bert} is chosen as the dialogue context encoder.

\paragraph{Grid World State Encoder.}
\label{sec:world_encoder}
The world state is represented by a voxel-based grid. % and is fed into the world state encoder at each time step.
We first represent each grid state as a 7-dimensional one-hot vector that stands for empty or a block having one of six colours, yielding a 7×11×9×11 world state representation. 

The structure of the world state encoder is similar to \citet{jayannavar2020learning}'s, that is, consisting of $k$ 3D-convolutional layers ($f_{1}$) with kernel size 3, stride 1 and padding 1, followed by a ReLU activation function. 
Between every successive pair of these layers there is a 1×1×1 3D-convolutional layer ($f_{2}$) with stride 1 and no padding followed by ReLU:
\begin{align}%
    % \footnotesize
    & W_{i} = \text{ReLU}(f^{i}_{2}(\text{ReLU}(f^{i}_{1}(W_{i-1})))), \\
    & W_{k} = \text{ReLU}(f^{i}_{1}(W_{k-1})),       
\end{align}
where $i = 1, 2, \dots , k-1$. $W_{k} \in \mathbb{R}^{d_c \times 11 \times 9 \times 11}$ is the learned world grid-based representation where $d_c$ is the dimension of each grid representation. Then we reshape $W_{k}$ into $W^{'} \in \mathbb{R}^{d_c \times 1089}$.

\begin{figure}[!t]
  \centering
  \includegraphics[width=0.45\textwidth]{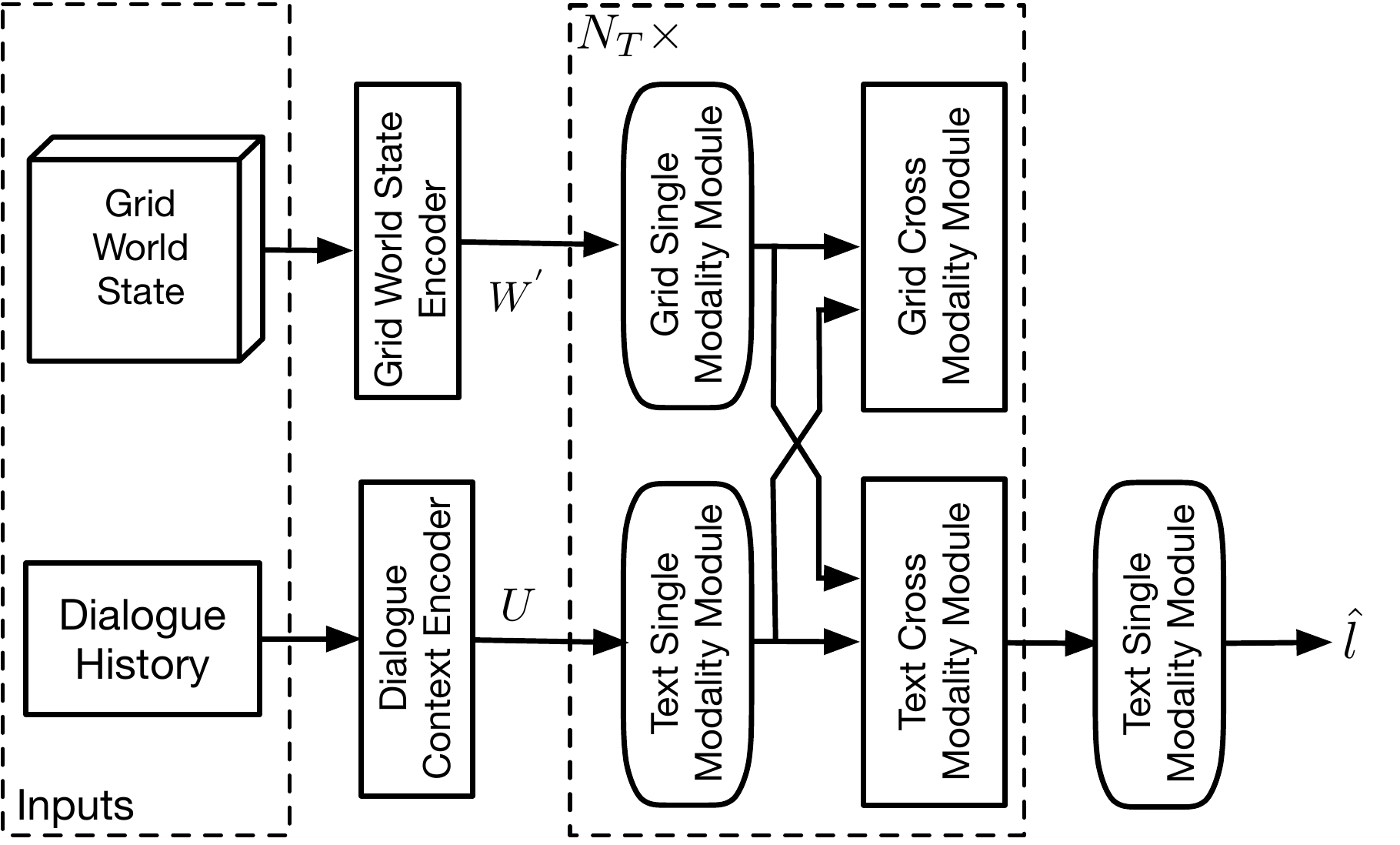}
  \caption{Model Architecture for the classification task.} %We propose an approach for LE extraction in this paper.}
  \label{fig:model}
\end{figure}

\paragraph{Fusion Module.}
\label{sec:fusin_module}
The fusion module comprises four major components: one \textit{single modality modules} and two \textit{cross-modality modules}. The former modules are based on self-attention layers and the latter on cross-attention layers. These take as input the world state representation and dialogue history representation. 
Between every successive pair of grid single-modality modules or text single-modality modules, there is a cross-modality module. We take $N_T$ layers for cross-modality modules.
We first revisit the definition and notations about the attention mechanism~\cite{bahdanau2014neural} and then introduce how they are integrated into our single-modality modules and cross-modality modules.
\paragraph{Attention Mechanism.}
Given a query vector $x$ and a sequence of context vectors $\{y_j\}^{K}_{j=1}$, the attention mechanism first computes the matching score $s_j$ between the query vector $x$ and each context vector $y_j$. Then, the attention weights are calculated by normalizing the matching score: %
$a_j = \frac{exp(s_j)}{\sum_{j=1}^{K}exp(s_j)}$. %
The output of an attention layer is the attention-weighted sum of the context vectors: %
$Attention(x, {y_j}) = \sum_{j} a_j \cdot y_j$. %
Particularly, the attention mechanism is called self-attention when the query vector itself is in the context vectors $\{y_j\}$. We use the multi-head attention following~\cite {devlin2018bert}.
\paragraph{Single-Modality Module.} 
Each layer in a single-modality module contains a self-attention sub-layer and a feed-forward sub-layer, where the feed-forward sub-layer is further composed of a linear transformation layer, a dropout layer and a normalization layer. 
We take $N_T$ and $N_T+1$ layers for the grid single-modality modules and the text single-modality modules respectively, interspersed with cross-modality module as shown in Figure \ref{fig:model}. Since new blocks can only be feasibly placed if one of their faces touches the ground or another block in the Minecraft world, we add masks to all infeasible grids in the grid single-modality modules. For a set of text vectors $\{u_i^{n}\}_{i=1}^{s}$ and a set of grid vectors $\{w_j^{m}\}_{j=1}^{1089}$ as inputs of $n$-th text single-modality layer and $m$-th grid single-modality layer, where $n\in \{1,\dots,N_T+1\}$ and $m \in \{1,\dots,N_G+1\}$, we first feed them into two self attention sub-layers:
\begin{align}%
    % \footnotesize
    % \small
    u_i^{n} &= \text{SelfAttn}_u^{n}(u_i^{n}, \{u_i^{n}\}), \\
    w_j^{m} &= \text{SelfAttn}_w^{m}(w_j^{m}, \{w_j^{m}\}, mask)
\end{align}
Lastly, the outputs of self attention modules, $u_i^{n}$ and $w_j^{m}$, are followed by feed-forward sub-layers to obtain $\hat{u_i}^{n}$ and $\hat{w_j}^{m}$.
\paragraph{Cross-Modality Module.} 
Each layer in the cross-modality module consists of one cross-attention sub-layer and one feed-forward sub-layer, where the feed-forward sub-layers follow the same setting as the single-modality module. 
Given the outputs of $n$-th text single-modality layer, $\{\hat{u_i}^{n}\}_{i=1}^{s}$, and the $m$-th grid single-modality layer, $\{\hat{w_j}^{m}\}_{j=1}^{1089}$, as the query and context vectors, we pass them through cross-attention sub-layers, respectively:
\begin{align}%
    % \footnotesize
    \hat{u_i}^{n+1} &= \text{CrossAttn}_u^{n}(\hat{u_i}^{n}, \{\hat{w_j}^{m}\}), \\
    \hat{w_j}^{m+1} &= \text{CrossAttn}_w^{m}(\hat{w_j}^{m}, \{\hat{u_i}^{n}\}),
\end{align}
The cross-attention sub-layer is used to exchange the information and align the entities between the two modalities in order to learn joint cross-modality representations. Then the output of the cross-attention sub-layer is processed by one feed-forward sub-layer to obtain $\{u_i^{n+1}\}_{i=1}^{s}$ and $\{w_j^{m+1}\}_{j=1}^{1089}$, which will be passed to the following singe-modality modules.

Finally, we obtain a set of word vectors, $\{\hat{u_i}^{N_T+1}\}_{i=1}^{s}$, and a set of grid vectors, $\{\hat{w_j}^{N_T}\}_{j=1}^{1089}$. We take word representations as $U^{N_T}$.

\paragraph{Slot Decoder.}
\label{sec:decoder}
The Slot Decoder contains one linear projection layers of trainable parameters $W$. We compute the average of $U^{N_T} \in \mathbb{R}^{s \times d_w}$ alongside the $s$-dimension to obtain $u \in \mathbb{R}^{d_w}$.
Then we obtain a scalar value for the binary classification via: 
$\hat{l} = \text{Sigmoid}(W \cdot u)$. We train the model by minimizing the cross-entropy loss.

\subsection{Results}
Table \ref{table:results} presents the performance of all comparison approaches for the classification task. 
Our proposed method achieves a 75.7\% $F_1$ score and secures the 3rd position among all the compared approaches.
It is important to note that the performance difference between the teams is relatively small.

\definecolor{legend1}{HTML}{66c2a4}
\definecolor{legend2}{HTML}{fa8c62}
\definecolor{legend3}{HTML}{8da0cb}
\definecolor{cid}{HTML}{dae8f5}
\definecolor{ccon}{HTML}{fee9d4}
\definecolor{gred}{HTML}{cc0200}
\definecolor{ggreen}{HTML}{4C9F26}
\definecolor{Gray}{gray}{0.93}
\newcommand\hl{\cellcolor{cid}}
\newcommand\se{\cellcolor{ccon}}

\begin{table*}[!ht]
\centering
\footnotesize
\begin{tabular}{lcccc}
\toprule
\bf Team Name  & \bf Rank     & \bf F1 Score (Nearest Bin)       &\bf F1 Score      &\bf  Successful Entries \\
\midrule
try1try & 1          & 0.750	& 0.766	 & 39   \\
felipe\_B & 2          & 0.750	& 0.761	 & 54   \\
Ours \hl& 3      \hl    & 0.750\hl	& 0.757\hl	 & 39 \hl  \\
testa & 4          & 0.750	& 0.756	 & 29   \\
745H1N & 5          & 0.750	& 0.756	 & 14   \\
ITL-ed & 6          & 0.750	& 0.754	 & 66   \\
craftsmanfly & 7          & 0.750	& 0.751	 & 74   \\
ITL-ed & 8          & 0.750	& 0.754	 & 66   \\
\bottomrule
\end{tabular}
\caption{Test Results from the IGLU NLP Task Leaderboard. Successful Entries stand for the number of trials. Our proposed method is highlighted in blue.}
\label{table:results}
\vspace{-1.5em}
\end{table*}

%%%%%%%%%%%%%%%%%%%%%%%%%%%
% Improvements on Ranking %
%%%%%%%%%%%%%%%%%%%%%%%%%%%

\section{Ranking Task}
In this section, we introduce our proposed method for the ranking task. We have made several efforts on improving the ranking task, including expanding the classical ranking methods, query expansion, and neural approaches. Details are as follows.

\subsection{Grid Search of BM25}
First, to find the best parameter of the BM25 ranker, we conducted a grid search of the two parameter \texttt{k1} and \texttt{b}.
From the grid search, we have found that \texttt{k1=1.8} and \texttt{b=0.98} can perform the best in the validation set.

% See Appendix listing \ref{lst:grid-search} for the logic of the grid search. The full code can be found in \tcbox{models/baseline-bert-classifier-bm25-ranker.ipynb}.

\subsection{Use of Classic Ranking methods with heuristic}
Next, we have explored multiple classic ranking techniques from Information Retrieval, including the TF-IDF, PL2, DPH weighting models as well as reproducing the results of BM25.

We implemented additional preprocessing by changing all characters to lowercase and removing punctuation. In addition, we added a \textbf{heuristic} for all clarification questions that scored a BM25 score of zero. We ranked all questions with a score of zero by the number of words, in ascending order. The reasoning is that shorter questions are more general and are able to answer a wider variety of questions, thus making it more likely to be the correct answer. Furthermore, long questions are more likely to contain more unique words, which means that it is highly unlikely to be the correct answer if none of the words matched the query. We found that these two steps improved the performance of the ranking task, and the results were as follows:
\begin{center}
\begin{BVerbatim}
MRR5:  0.36846441947565517
MRR10: 0.38006777242732276
MRR20: 0.388398839553438
\end{BVerbatim}
\end{center}
% See Appendix listing \ref{lst:ranking-heuristic} for the logic of the heuristic.

\subsection{Query Expansion}
In order to improve the ranking performance, we experimented with a more recent approach of ranking techniques.
Query expansion techniques have shown their advantages in improving the recall performance of a ranking model. In our experiments, we explored the use of the Bo1, KL divergence-based and RM3 query expansion techniques in a re-ranking setup with a TF-IDF weighting model. In addition, we also extend the  query expansion technique with a fine-tuned T5 model. A T5 model is trained on instruction-clarification question pairs and we aim to automatically generate possible clarification questions to be added to a given instruction (i.e., query) for query expansion and then leverage various classic weighting models (TF-IDF, BM25, PL2, DPH) for clarification question (document) retrieval.

\subsection{Reranking with Semantic Textual Similarity (Semantic Search)}
We used a Sentence Transformer's RoBERTa model as a sentence encoder to get a Semantic Textual Similarity (STS) between instructions and clarifying questions. We encoded instructions and clarifying questions separately but with the same encoder. Then, the STS could be computed with the cosine similarity between the embedding of instruction and question.
After we first rank the questions with the BM25, we reranked the top 20 of the result with the score of STS.
% Code can be found in \tcbox{models/rankers/reranker\_sts.py} and \tcbox{models/BERT-bm25-rerankSTS.ipynb}.

\subsection{Reranking with BERT model}
We attempted to do reranking using BERT and RoBERTa models. We encoded the instruction input and clarification question and trained a classifier to calculate the probability that the instruction input matches the clarification question. We then reranked the clarification questions collected by BM25 by the highest probability. Given the high skew towards negative samples, we decided to have the dataset contain 33\% positive labels and 66\% negative labels. However, since the size of the dataset is small, we found it difficult to train a reranker model that significantly improved ranking performance.
% Code can be found in \tcbox{models/ReRankerClassification.ipynb}.

\subsection{Use of Bi-encoder and Cross-encoder}
We attempted various combinations of ranker and reranker.
First, we tried using a bi-encoder only for the ranking task. The bi-encoder model is \textit{sentence-transformers/multi-qa-MiniLM-L6-cos-v1}\footnote{https://huggingface.co/sentence-transformers}, which was trained for semantic search.

The ranking results were like the following:
\begin{center}
\begin{BVerbatim}
MRR5:  0.2026591760299625
MRR10: 0.21386614945603696
MRR20: 0.2222604358213725
\end{BVerbatim}
\end{center}

Next, we added a cross-encoder for the reranker. Two models we have tried are \textit{cross-encoder/ms-marco-MiniLM-L-12-v2} and \textit{cross-encoder/ms-marco-MiniLM-L-6-v2}\footnote{https://huggingface.co/cross-encoder}, which are both trained on the MS Marco Passage Ranking task.
The ranking results were improved compared to the previous attempt.
\begin{center}
\begin{BVerbatim}
MRR5:  0.26166666666666677
MRR10: 0.27329498840734795
MRR20: 0.28037113927919943
\end{BVerbatim}
\end{center}

Since BM25 is still a great retriever, we used BM25 as an initial ranker and use a cross-encoder as reranker.
The score was improved as follows:
\begin{center}
\begin{BVerbatim}
MRR5:  0.2841760299625468
MRR10: 0.2980809702158013
MRR20: 0.30612438951917825
\end{BVerbatim}
\end{center}

% Code can be found in \tcbox{models/BERT-biencoder-crossencoder.ipynb}.

\section{Discussion and Future Direction}
%% Future for classification task
\paragraph{Classification Task.}
To further enhance the model performance, it may be beneficial to explore larger language models \cite{kaplan2020scaling,hoffmann2022training}, prompting-oriented methods \cite{schick-schutze-2021-exploiting}, or further pre-training \cite{gururangan-etal-2020-dont,shi2023don} in future research.

%% Future for retrieval and ranking
\paragraph{Ranking Task.}
Although we have experimented with recent approaches, the BM25 with added heuristic performed the best. We will have to research more about the reason behind it in the future.
In the literature, advanced dense retrieval approaches have also been experimented with, including MonoT5 and ColBERT. We plan to continue our research with these models on the IGLU competition dataset.

%% ClariQ and Instruction Generation - Clarifying Question (ClariQ) and Instruction Generation
\paragraph{Generation of Instructions and Clarifying Questions.}
As an additional task, we can try to generate pairs of instruction and clarifying questions.
The generation of those can help augment the training data for the classification and ranking models.

Not only that, this direction aligns with the recent research in user simulation \cite{kim-2022} and Theory of Mind \cite{neural-tom}, where they view conversation as a cooperative and collaborative mind game based on the study of \textit{pragmatics}. 
According to this view, it is important for a conversational system to be able to model users (listeners) \cite{fried-etal-2018-pragmatics}, i.e., endowing a conversational system with an ability to reason counterpart’s mental states and realities of the surroundings.
With this aspect, clarifying question generators can act as a user (Builder) simulator in the interactive conversational environment, enabling the Architect to generate instructions that are less ambiguous to the Builder.
Moreover, it is worth exploring if we can find a meaningful relationship between the satisfaction level of the Builder and the need for clarifying questions. Estimating the Builder's satisfaction level, following the work of \citet{kim-2022}, may help classify the need for clarifying questions.

%%%%%%%%%%%%%%%%%
%.  References  %
\clearpage
\bibliography{references}
\bibliographystyle{acl_natbib}

%%%%%%%%%%%%%%%
%.  Appendix  %
%%%%%%%%%%%%%%%

% Change to the single-column style for code listing in the appendix
% \onecolumn
% \appendix

% \section{Code Snippets}\label{appendix:code}

% \begin{lstlisting}[language=Python, label={lst:ranking-heuristic}, caption=Ranking Heuristic]
% bm25_ascending = sorted(bm25_corpus, key=len)
% cntr = 0
% while len(bm25_ranked_list) < top_k and cntr < len(bm25_ascending):
%     if bm25_ascending[cntr] not in bm25_ranked_list:
%         bm25_ranked_list.append(bm25_ascending[cntr])
%     cntr += 1
% \end{lstlisting}

% \begin{lstlisting}[language=Python, label={lst:grid-search}, caption=Grid Search of BM25 Hyperparameters]
% best_k1, best_b = None, None
% for k1 in range(11):
%     k1 = 2.0 + k1 / 10
%     for b in range(11):
%         b = 0.9 + b / 100
% \end{lstlisting}

\end{document}